\documentclass[10pt,twocolumn,letterpaper]{article}

\usepackage{3dv}
\usepackage{times}
\usepackage{epsfig}
\usepackage{graphicx}
\usepackage{amsmath}
\usepackage{amssymb}
\usepackage{comment}
\usepackage{color}

\usepackage[pagebackref=true,breaklinks=true,letterpaper=true,colorlinks,bookmarks=false]{hyperref}

\makeatletter
\DeclareRobustCommand\onedot{\futurelet\@let@token\@onedot}
\def\@onedot{\ifx\@let@token.\else.\null\fi\xspace}

\usepackage{multirow}
\usepackage{epsfig}
\usepackage{subcaption}
\def\eg{\emph{e.g}\onedot} 
\def\ie{\emph{i.e}\onedot}

\def\wrt{w.r.t\onedot}

\newcommand{\trans}{\text{T}}
\newcommand{\matr}[1]{\mathbf{#1}}     

\newcommand{\Proj}{\matr{P}}
\newcommand{\Hom}{\matr{H}}

\newcommand{\Intrinsic}{\matr{K}}

\threedvfinalcopy 

\def\threedvPaperID{215} 

\ifthreedvfinal\fi
\begin{document}

\title{Pose Estimation for Vehicle-mounted Cameras via\\Horizontal and Vertical Planes}

\author{Istan Gergo Gal\\
Department of Algorithms and their Applications\\
Eotvos Lorand University, Budapest, Hungary
\and
Daniel Barath\\
VRG, Department of Cybernetics, Czech Technical University in Prague, Czech Republic \\
and Machine Perception Research Laboratory, MTA SZTAKI, Budapest, Hungary 
\and
Levente Hajder\\
Department of Algorithms and their Applications\\
Eotvos Lorand University, Budapest, Hungary
}

\maketitle

\begin{abstract}
We propose two novel solvers for estimating the egomotion of a calibrated camera mounted to a moving vehicle from a single affine correspondence via recovering special homographies. For the first class of solvers, the sought plane is expected to be perpendicular to one of the camera axes. For the second class, the plane is orthogonal to the ground with unknown normal, \eg, it is a building facade.
Both methods are solved via a linear system with a small coefficient matrix, thus, being extremely efficient. Both the minimal and over-determined cases can be solved by the proposed methods. They are tested on synthetic data and on publicly available real-world datasets. The novel methods are more accurate or comparable to the traditional algorithms and are faster when included in state of the art robust estimators.
\end{abstract}

\section{Introduction}
The estimation of plane-to-plane correspondences (\ie, homographies) in an image pair is a fundamental problem for recovering the scene geometry or the relative motion of the camera. 
Many state of the art (SOTA) Structure from Motion~\cite{snavely2006photo,snavely2008modeling,schonberger2016structure}  (SfM) or Simultaneous Localization and Mapping~\cite{durrant2006simultaneous,bailey2006simultaneous} (SLAM) algorithms combine epipolar geometry and homography estimation to be robust when the scene is close to being planar or the camera motion is rotation-only -- those cases when the epipolar geometry is not estimable.  
In this paper, we use non-traditional input data (\ie, affine correspondence) and focus on a special case when the camera is mounted to a moving vehicle and there is a prior knowledge about the sought plane, for instance, it is the ground or a building facade. 

Nowadays, a number of algorithms exist for estimating or approximating geometric models, \eg, homographies, using affine correspondences. An affine correspondence consists of a point pair and the related $2\times2$ local affine transformation mapping the infinitesimally close vicinity of the point in the first image to the second one.
A technique, proposed by Perdoch et al.~\cite{PerdochMC06}, approximates the epipolar geometry from one or two affine correspondences by converting them to point pairs.   
Bentolila and Francos~\cite{Bentolila2014} proposed a solution for estimating the fundamental matrix using three affine features. 
Raposo et al.~\cite{Raposo2016,raposo2016pi} and Barath et al.~\cite{barath2018efficient} showed that two correspondences are enough for estimating the relative camera motion. 
Moreover, two feature pairs are sufficient for solving the semi-calibrated case, \ie, when the objective is to find the essential matrix and a common unknown focal length~\cite{barath2017focal}. 
Furthermore, homographies can be estimated from two affine correspondences~\cite{koser2009geometric} and in case of known epipolar geometry, from a single correspondence~\cite{barath2017theory}. 
There is a one-to-one relationship between local affine transformations and surface normals~\cite{koser2009geometric,barath2015optimal} for calibrated image pairs. 
Pritts et al.~\cite{Pritts2017RadiallyDistortedCT} showed that the lens distortion parameters can be retrieved as well.
Affine features encode higher-order information about the scene geometry, thus the previously mentioned algorithms solve geometric estimation problems from fewer features than point-based methods. 

For obtaining affine correspondence, one can apply one of the traditional affine-covariant feature detectors, thoroughly surveyed in~\cite{mikolajczyk2005comparison}, such as MSER~\cite{MSER}, Hessian-Affine, Harris-Affine, etc. 
Another way of acquiring affine features is via view-synthesizing, \eg, as done in Affine-SIFT~\cite{Morel2009} or MODS~\cite{mishkin2015mods}, by sampling the affine space and affinely warping the input images. 
Moreover, there are modern learning-based approaches, \eg, the Hes-Aff-Net~\cite{mishkin2018repeatability} which obtains affine regions by running CNN-based shape regression on Hessian keypoints.

The attention recently is pointing towards autonomous driving thus, it is becoming more and more important to design algorithms exploiting the properties of such a movement to provide results superior to general solutions.
Considering that the cameras are moving on a plane, \eg, they are mounted to a car,  is a well-known approach for reducing the degrees-of-freedom, thus, speeding up the robust estimation.
Note that this assumption can be made valid if the vertical direction is known, \eg, from an IMU sensor.

Ortin and Montiel~\cite{Ortin2001} proved that, in case of planar motion, the epipolar geometry can be estimated from two point correspondences. 
Since then, several solvers have been proposed to estimate the motion from two correspondences~\cite{chou2015,choi2018}. 
Scaramuzza~\cite{Scaramuzza2011} proposed a technique using a single point pair for a special camera setting assuming the special non-holonomic constraint to hold.

The most related work to ours is the paper of Saurer et al.~\cite{saurer2016homography}. 
They estimate homographies from point correspondences by considering that there is a prior knowledge about the normal of the sought plane, \eg, it is orthogonal or parallel to the plane on which the vehicle, \ie, typically a (quad)copter, moves. In their paper, the camera rotates around the optical axis contrary to this work, where we assume planar motion.

\begin{figure}
  	\centering
  	\includegraphics[width=0.9\columnwidth]{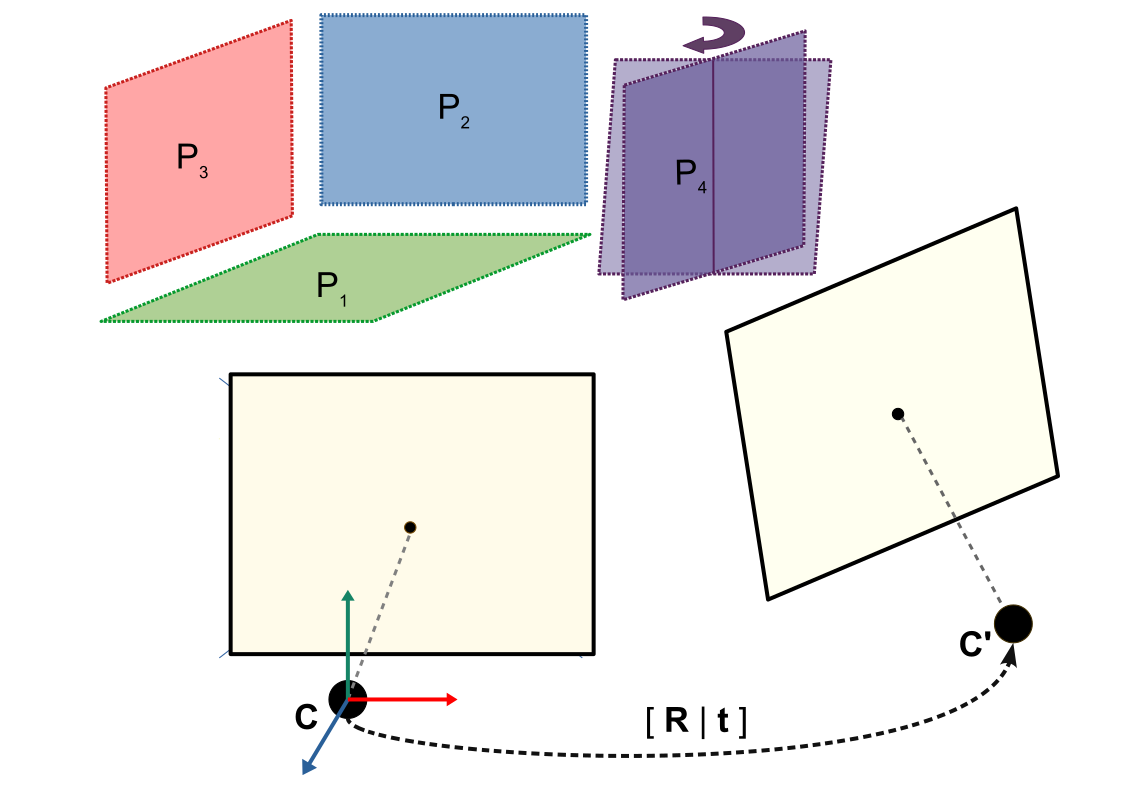}
  	\caption{Visualization of the configuration.  
  	Cameras $\matr C$ and $\matr C'$ related by a rotation $\matr R$ around the vertical (yaw) axis and translation $\matr t = [t_x, 0, t_z]^\trans$.
  	Four different cases are considered, \ie, when the points originate from a (1) horizontal $\text{P}_1$, (2) vertical frontal $\text{P}_2$, (3) vertical side $\text{P}_3$, and (4) general vertical plane $\text{P}_4$.}
	\label{fig:Tekla}
\end{figure}

\noindent \textbf{Contribution.}
We propose two classes of solvers for \textit{estimating} special homographies \textit{from a single affine correspondence}. Planar motion is assumed here.
For the first one, the plane is assumed to be perpendicular to one of the camera axes.
For the second one, the plane is vertical, \eg, it is a facade of a building, with its normal represented by a single angle.
The proposed methods are solved as a linear system, thus, being extremely fast, \ie, $5$--$10$ $\mu$s.
The methods are tested on synthetic data and on publicly available real-world datasets. They lead to accuracy better than the traditional algorithms while being faster when included in SOTA robust estimators.

\section{Problem Statement}

Given two calibrated cameras, with intrinsic camera matrices $\Intrinsic$ and ${\Intrinsic}'$, a planar object is observed.
The world coordinate system is fixed to the first camera.
The projection matrices are $\Proj = \Intrinsic \left[ \matr I \, | \, \matr 0 \right]$, ${\Proj}' = {\Intrinsic}' \left[ \matr R \, | \, \matr t \right]$, 
where matrix $\matr R$ and vector $\matr t$ are, respectively, the 3D rotation and translation between the two views.

If there are corresponding points in the images, given by homogeneous coordinates as $\matr u=[x \quad y\quad 1]$ and $\matr u'=[x'\quad y' \quad 1]$, then the relationship \wrt the coordinates is linear. It is represented by a homography $\Hom$ as $\matr u' \sim \Hom \matr u$, where the operator $\sim$ denotes equality up to an usually unknown scale. 

In the case of calibrated cameras, the 2D coordinates can be normalized by the inverse of the intrinsic camera matrices. For the sake of simplicity, we use the normalized coordinates in the rest of this paper: $ \matr u \leftarrow \Intrinsic^{-1} \matr u $ and $\matr u' \leftarrow \Intrinsic'^{-1} \matr u'$.

The homography parameters can be expressed via the relative camera pose~\cite{hartley2003multiple}, \ie, rotation and translation, as follows:
\begin{equation}
\label{eq:basic_homography}
\Hom \sim \matr R - \frac{1}{d} \matr t \matr n^\trans,
\end{equation}
where scalar $d$ and vector $\matr n$ denote the distance of the observed plane from the first image and the normal of the plane, respectively. 

\subsection{Planar motion} 
Assume that it is given a calibrated image pair with a common $\text{XZ}$ plane ($\text{Y}=0$), where axis $\text{Y}$ is parallel to the vertical direction of the image planes. 
A trivial example for such constraint is the camera setting of autonomous cars with a camera fixed to the moving vehicle and the $\text{Y}$ axis of the camera being perpendicular to the ground plane. 
Note that this constraint can be straightforwardly made valid if the vertical direction is known, \eg, from an IMU sensor.
To estimate the camera motion, we first describe the parameterization of the problem.

\begin{figure}
  	\centering
  	\includegraphics[width=0.85\columnwidth]{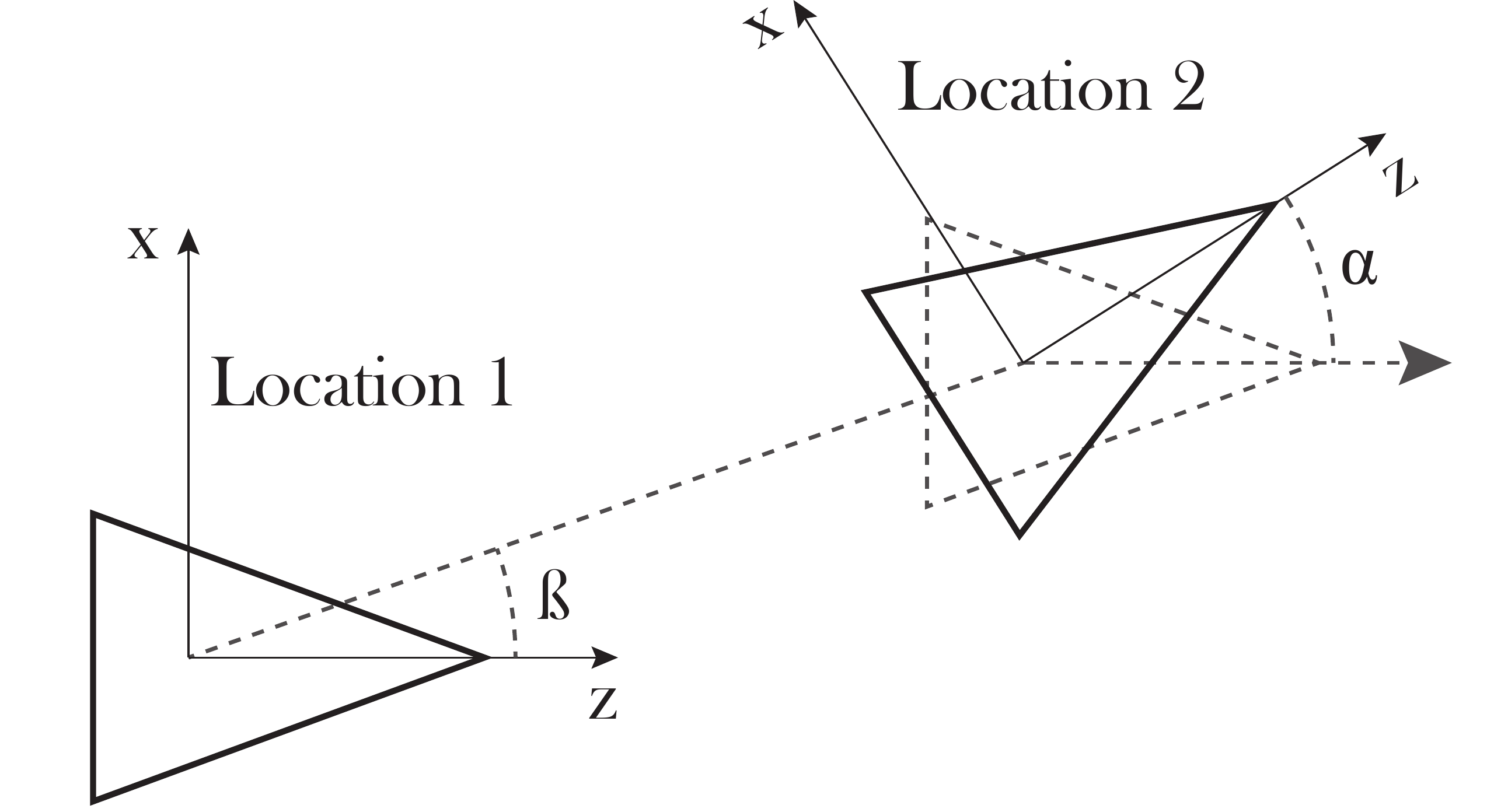}
  	\caption{Motion scheme. The camera movement is described by the angle $\alpha$ of the rotation perpendicular to axis $\text{Y}$ and translation vector $[\cos(\beta), \; 0 \;, \sin(\beta)]^\trans$. The length of the translation is not considered as it cannot be reconstructed from images.}
	\label{fig:illustration_image}
\end{figure}

Assuming planar motion, the rotation and translation are represented by three parameters: a 2D translation and the angle of rotation. Formally,
\begin{eqnarray}
\label{eq:planar}
\begin{array}{cc}
\matr{R} = \left[\begin{array}{ccc}
 \cos  \alpha & 0 & -\sin  \alpha \\
 0 & 1 & 0 \\
 \sin  \alpha & 0 &  \cos  \alpha \\
\end{array}\right],
&
\matr{t} = \rho 
\left[\begin{array}{ccc}
\cos \beta \\
0 \\
\sin \beta
\end{array}\right].
\end{array}
\end{eqnarray}
The 2D translation is represented by angle $\beta \in [0, 2 \pi)$ and length $\rho \in \matr R^{+}$. Angle $\alpha \in [0, 2 \pi)$ is the rotation around axis $\text{Y}$. The setup is shown in Fig~\ref{fig:illustration_image}. 

\subsection{Homography Estimation}

In this paper, we exploit the relation between homographies and local affine frames. A homography $\Hom$ is represented by a $3 \times 3$ and an affinity $\matr A$ by a $2 \times 2$ matrices as follows:
\begin{equation*}
\Hom =
	\begin{bmatrix}
	h_1 & h_2 & h_3 \\
	h_4 & h_5 & h_6 \\
	h_7 & h_8 & h_9 
	\end{bmatrix} ,  \quad \quad \quad
\matr A=
	\begin{bmatrix}
	a_1 & a_2 \\
	a_3 & a_4 \\
	\end{bmatrix}.
\end{equation*}
Homography $\Hom$ represents the projective transformation between corresponding areas of planar surfaces in two images, while affine transformations are defined as the first-order approximations of image-to-image  transformations~\cite{barath2015optimal}, including homographies.
A homography is usually estimated from point correspondences in the images. 
If the point locations are denoted by vector $[x\quad y]^\trans$ and $[x' \quad y']^\trans$ in the first and second images, the relations between the coordinates~\cite{hartley2003multiple} in the two views are as follows:
\begin{eqnarray}
\begin{array}{c}
    \label{eq:hom_pt}
    x' \left(h_7 x + h_8 y +h_9 \right)=h_1 x + h_2 y +h_3,  \\
    y' \left(h_7 x + h_8 y +h_9 \right)=h_4 x + h_5 y +h_6.
    \end{array}
\end{eqnarray}
Thus, each point correspondence (PC) adds two equations for the homography estimation.

Recently, Barath and Hajder proved~\cite{barath2017theory} that the affine part of an affine correspondence gives four additional equations. They are as follows: 
\begin{eqnarray}
\label{eq:hom_affine}
\begin{array}{c}
    h_1 - \left(x'+a_1 x \right) h_7 - a_1 y h_8 - a_1 h_9 = 0,  \\ 
    h_2 - \left(x'+a_2 y \right) h_8 - a_2 x h_7 - a_2 h_9 = 0, \\
    h_4 - \left(y'+a_3 x \right) h_7 - a_3 y h_8 - a_3 h_9 = 0,  \\
    h_5 - \left(y'+a_4 y \right) h_8 - a_4 x h_7 - a_4 h_9 = 0. 
\end{array}
\end{eqnarray}
In total, an affine correspondence (AC) provides six independent constraints.
Consequently, one AC and one PC are enough for estimating a general homography with 8 degrees-of-freedom. 

\section{Proposed Methods}
The following three problem classes are considered: the estimation of (i) the ground plane, (ii) special vertical planes, (iii) general vertical planes. The main objective is to recover the camera pose and the surface normal.

\subsection{Ground Plane}
The normal of the ground plane\footnote{Plane $\text{P}_\text{1}$ of Fig.~\ref{fig:Tekla}} is 
$	\matr n = 	\begin{bmatrix} 0 & 1 & 0  \end{bmatrix} ^\trans$. If planar motion is considered, Eq.~\ref{eq:planar} and normal $\matr n$ are substituted into Eq.~\ref{eq:basic_homography}. 
The homography matrix can be written as follows:
	\begin{equation*}
	\gamma \Hom = 	
	\begin{bmatrix}
	\cos \alpha &  0 & -\sin \alpha  \\
	0 & 1 & 0\\ 
	\sin \alpha  & 0 & \cos \alpha
	\end{bmatrix}
	-
	\rho \begin{bmatrix}
	\cos \beta \\ 0 \\ \sin \beta 
	\end{bmatrix}
	\begin{bmatrix}
	0 \\ 1 \\ 0
	\end{bmatrix}^\trans,
	\end{equation*}
where parameter $\gamma$ denotes the unknown scale. If one expresses the nine elements of the homography, the following formula is obtained:
	\begin{eqnarray}
\begin{array}{cc}
	\gamma \Hom  = 
	\begin{bmatrix}
	\cos \alpha  &  p & -\sin \alpha \\
	0 & 1 & 0\\ 
	\sin\alpha & q & \cos\alpha
	\end{bmatrix},
 \\
 \\
		\Hom \sim 
	\begin{bmatrix}
	h_{1} & h_{2} & h_{3}\\
	0 & h_{5} & 0 \\
	h_{7} & h_{8} & h_{9}
	\end{bmatrix},
	\label{eq:HSikMozgasVizszint}
	\end{array}
	\end{eqnarray}
where  $p=-\rho \cos \beta$ and $q=-\rho \sin \beta$, and the elements of $\Hom$ are $h_1=h_9=\cos \alpha$,  $h_2=p$, $h_3=-h_7=-\sin \alpha$, $h_8=q$, $h_5=1$. Accordingly, three degrees-of-freedom (DoF) have to be estimated, \ie, the unknown rotation angle $\alpha$ and a 2D direction represented by coordinates $p$ and $q$. The latter ones encode the movement of the second camera \wrt the first one.

If the relationship of the homography, point and affine parameters are considered (Eqs.~\ref{eq:hom_pt} and~\ref{eq:hom_affine}), the estimation problem can be linearized as follows:
\begin{equation}
\matr A_{gd}
\left[\begin{array}{c}

\cos\alpha\\
\sin\alpha\\
p\\
q
\end{array}\right]=\left[\begin{array}{cccccc}
0 &
-y &
0 &
0 &
0 &
-1
\end{array}\right]^\trans,
\label{eq:ground}
\end{equation}
where:
\begin{equation*}
\scriptsize
\matr A_{gd} = \left[\begin{array}{cccc}
x-x' & -x'x-1 & y & -x'y\\
-y' & -y'x & 0 & -y'y\\
1-a_{1} & -x'-a_{1}x & 0 & -a_{1}y\\
-a_{2} & -a_{2}x & 1 & -x'-a_{2}y\\
-a_{3} & -y'-a_{3}x & 0 & -a_{3}y\\
-a_{4} & -a_{4}x & 0 & -y'-a_{4}y
\end{array}\right].
\end{equation*}
The point and affine parameters give six linear equations in the form of $\matr A_{gd} \matr h_{gd}=\matr b_{gd}$. The problem can be written as $\matr A_{gd} \matr x = \matr b_{gd}$.

\noindent \textbf{Optimal solver.} 
The objective is to solve an inhomogeneous linear system with constraint $x^2_1 + x^2_2=1$, where $x_1=\cos \alpha$ and  $x_2=\sin \alpha$ are the first two coordinates of vector $\matr x$. 
As it is proven in the supplementary material, this algebraic problem can be optimally solved in the least squares sense by computing the intersections of two conics.  

\noindent \textbf{Rapid solver.} Although the optimality is lost, the problem can be solved by a homogeneous linear equation if reformulated as $\left[\matr A_{gd}|-\matr b_{gd} \right] [\begin{array}{cc} \matr x^\trans & 1 \end{array}]^\trans= \matr{0}$.
The null-vector of matrix $\left[\matr A_{gd}|-\matr b_{gd} \right]$ gives the suboptimal solution.
Constraint $x^2_1 + x^2_2=1$ is made valid by dividing the obtained vector by its last coordinate. 

The angle itself can be retrieved by $\alpha=atan2(x_2,x_1)$.

\subsection{Special Vertical Planes}
For urban scenes, it is quite frequent that planes of the buildings are parallel or perpendicular to the moving direction of the vehicle.\footnote{Planes $\text{P}_\text{2}$ and $\text{P}_\text{3}$ of Fig.~\ref{fig:Tekla}} In these cases, normals are $[1 \quad 0 \quad 0]^\trans$ or $[0 \quad 0 \quad 1]^\trans$. The related homography matrices are written as 
\begin{equation*}
	\matr \Hom_{s_1} \sim 
	\begin{bmatrix}
	\cos \alpha -p  &  0 & -\sin \alpha \\
	0 & 1 & 0\\ 
	\sin\alpha -q & 0 & \cos\alpha
	\end{bmatrix},
	\end{equation*}
	\begin{equation*}
	\matr H_{s_2} \sim
	\begin{bmatrix}
	\cos \alpha  & 0  & -\sin \alpha -p \\
	0 & 1 & 0\\ 
	\sin\alpha & 0 & \cos\alpha -q
	\end{bmatrix}.
\end{equation*}
Although the homography matrices are not exactly the same as in Eq.~\ref{eq:HSikMozgasVizszint}, the problem is linear \wrt the same unknown parameters $\alpha$, $p$ and $p$.  Therefore, the problem can be solved straightforwardly, thus the same solver can be applied, only the coefficient matrices have to be modified. They are as follows:
\begin{eqnarray}
\scriptsize
\begin{array}{c}
\matr A_{v_1}=\left[\begin{array}{cccc}
x-x' & -1-x'x & -x & x'x\\
-y' & -y'x & 0 & y'x\\
1-a_{1} & -a_{1}x-x' & -1 & a_{1}x+x'\\
-a_{2} & -a_{2}x & 0 & a_{2}x\\
-a_{3} & -a_{3}x-y' & 0 & a_{3}x+y'\\
-a_{4} & -a_{4}x & 0 & a_{4}x
\end{array}\right],
\\
\\
\matr A_{v_2}=\left[\begin{array}{cccc}
x-x' & -1-x'x & -1 & x'\\
-y' & -y'x & 0 & y'\\
1-a_{1} & -a_{1}x-x' & 0 & a_{1}\\
-a_{2} & -a_{2}x & 0 & a_{2}\\
-a_{3} & -a_{3}x-y' & 0 & a_{3}\\
-a_{4} & -a_{4}x & 0 & a_{4}
\end{array}\right].
\end{array}
\label{eqs:verical}
\end{eqnarray}
The algebraic problems can be written as
\begin{eqnarray*}
\begin{array}{c}
\matr A_{v_1}\left[\cos \alpha,\sin \alpha, p, q \right]^\trans = \matr b_{v_1},\\
\matr A_{v_2}\left[\cos \alpha,\sin \alpha, p, q \right]^\trans = \matr b_{v_2}, 
\end{array}
\end{eqnarray*}
where the right sides of the inhomogeneous problems are exactly the same as in Eq.~\ref{eq:ground}, thus $\matr b_{v_1}=\matr b_{v_2}=\matr b_{gd}$.   

\subsection{General Vertical Planes}

Assuming that the observed plane is vertical, with unknown orientation, is also an important case for autonomous driving.\footnote{Plane(s) $\text{P}_\text{4}$ of Fig.~\ref{fig:Tekla}}
A general vertical wall has normal $\matr n=[n_x \quad 0 \quad n_z]^\trans$. The surface normal itself can be represented by an angle $\delta$ as $\matr n=[\cos \delta \quad 0 \quad \sin \delta]^\trans$.
The implied homography is as follows: 
\begin{equation}
\gamma \Hom =
	\begin{bmatrix}
	\cos \alpha-p \cos \delta &  0 & \sin \alpha -p \sin \delta \\
	0 & 1 & 0\\ 
	-\sin \alpha - q \cos \delta & 0 & \cos \alpha- q \sin \delta
	\end{bmatrix}.
\label{eq:genVertPlane}
\end{equation}
The equation can be written as:
\begin{equation*}
   	\Hom = 
	\begin{bmatrix}
	h_{1} & 0 & h_{3}\\
	0 & h_5 & 0 \\
	h_{7} & 0 & h_{9}
	\end{bmatrix}, 
\end{equation*}

where $h_1=(\cos \alpha-p \cos \delta)/\gamma$, $h_3=(\sin \alpha-p \sin \delta)/\gamma$, $h_5=1/\gamma$, $h_7=(-\sin \alpha-q \cos \delta)/\gamma$, and $h_9=(\cos \alpha-q \sin \delta)/\gamma$. Therefore, the problem has five DoF, \ie, $\alpha$, $\delta$, $\gamma$, $p$ and $q$.

\noindent The six linear equations from Eqs.~\ref{eq:hom_pt} and~\ref{eq:hom_affine} form the following linear system:
\begin{equation}
\scriptsize
	\begin{bmatrix}
    1  & 0 & 0 & -(x' + a_{1}x)  & -a_{1}\\
    0  & 0 & 0 & -a_{2}x  & -a_{2}\\
    0  & 0 & 0 & -(y' + a_{3}x)  & -a_{3}\\
    0  & 0 & 1 & -a_{4}x  & -a_{4}\\
    x  & 1 & 0 & -xx' & -x' \\
    0  & 0 & y & -xy' & -y' \\
	\end{bmatrix}
	\begin{bmatrix}
	h_1 \\ h_3 \\ h_5 \\ h_7 \\ h_9
	\end{bmatrix}
	= 
	\matr 0,
\end{equation}
or $\matr A_{vert} \matr h=0$ in short.

\noindent \textbf{Solver.} The elements of the homography matrix can be estimated by the null-matrix of $\matr A_{vert}$, and the scale-ambiguity, represented by variable $\gamma$, can be eliminated by scaling the homography matrix as $h_5=1$.

The remaining four parameters are retrieved from the scaled homography matrix.  The elements are written as:
\begin{equation}
\label{eq:hom_elements}
\begin{array}{c}
h_{1}=\cos\alpha-p\cos \delta, \\
h_{3}=\sin\alpha-p\sin \delta, \\
h_{7}=-\sin\alpha-q\cos \delta, \\
h_{9}=\cos\alpha-q\sin \delta.
\end{array}
\end{equation}
From the first and third equations parameters $p$ and $q$ can be expressed as:
\begin{equation}
\label{eq:p_and_q}
\begin{array}{cc}
p=\frac{\cos\alpha-h_{1}}{\cos \delta}, &
q=-\frac{h_{7}+\sin\alpha}{\cos \delta}.
\end{array}
\end{equation}
These are substituted back to the second and fourth equations. After elementary modifications, the following two equations are obtained:
\begin{eqnarray*}
\begin{array}{c}
h_{3}\cos\beta = h_{1}\sin\delta+\sin\left(\alpha-\delta\right),\\
h_{9}\cos\delta = h_{7}\sin\delta+\cos\left(\alpha-\delta\right).
\end{array}
\end{eqnarray*}
This can written by a matrix-vector product as:
\begin{equation*}
\left[\begin{array}{cc}
h_{9} & -h_{7}\\
h_{3} & -h_{1}
\end{array}\right]\left[\begin{array}{c}
\cos\delta\\
\sin\delta
\end{array}\right]=\left[\begin{array}{c}
\cos\left(\alpha-\delta\right)\\
\sin\left(\alpha-\delta\right)
\end{array}\right],
\end{equation*}
that is a constrained matrix-vector equation: $ \matr B \matr v_1 = \matr v_2$
\emph{s.t.} $\matr v_1^\trans \matr v_1 =\matr v_2^\trans \matr v_2 = 1$.

The SVD decomposition of matrix $\matr B$ is
$
    \matr B = \matr R_1
    diag(\sigma_1^2,\sigma_2^2)
    \matr R_2
$, where $\matr R_1$ and $\matr R_2$ are orthonormal matrices, \ie, rotations in 2D. Then the algebraic problem is as follows:
\begin{equation*}
    \matr B \matr v_1 = \matr R_1
    \begin{bmatrix}
    \sigma_1^2 &0\\
    0 & \sigma_2^2
    \end{bmatrix}
    \matr R_2 \matr v_1
    = \matr v_2.
\end{equation*}
\\
Multiplying both sides by $\matr R_1^\trans$ gives the formulas 
$
    diag(\sigma_1^2,\sigma_2^2)
    \matr R_2 \matr v_1
    = \matr R_1 ^\trans \matr v_2
$, where $\matr v_1^{'} = \matr R_2  \matr v_1 $ and $\matr v_2^{'} = \matr R_1^\trans \matr v_2 $.
\noindent Finally, the formula to be solved is:
\begin{equation}
    \begin{bmatrix}
    \sigma_1^2 &0\\
    0 & \sigma_2^2
    \end{bmatrix}
    \matr v_1^{'} 
    = \matr v_2^{'} .
\end{equation}

\noindent Note that $\matr v_1^{'\trans} \matr v_1^{'}$ = $\matr v_2^{'\trans} \matr v_2^{'} = 1$  since a rotation does not change the length of
a vector. This is a simple geometric problem: an origin-centered ellipse and an origin-centered circle with unit radius are on the left and right side of the equation, respectively. The intersections give four candidate solutions. 
The solution is straightforward and described in the supplementary material in depth. 

From the candidate solutions, the good one can be selected by the standard cheirality test built on the fact that all 3D points, from which the pose is calculated, should be located in front of both cameras.


\section{Experimental Results}

\begin{figure*}[t!]
  	\centering
	\begin{subfigure}[h]{0.30\textwidth}
	    \centering
          	\includegraphics[width=\columnwidth]{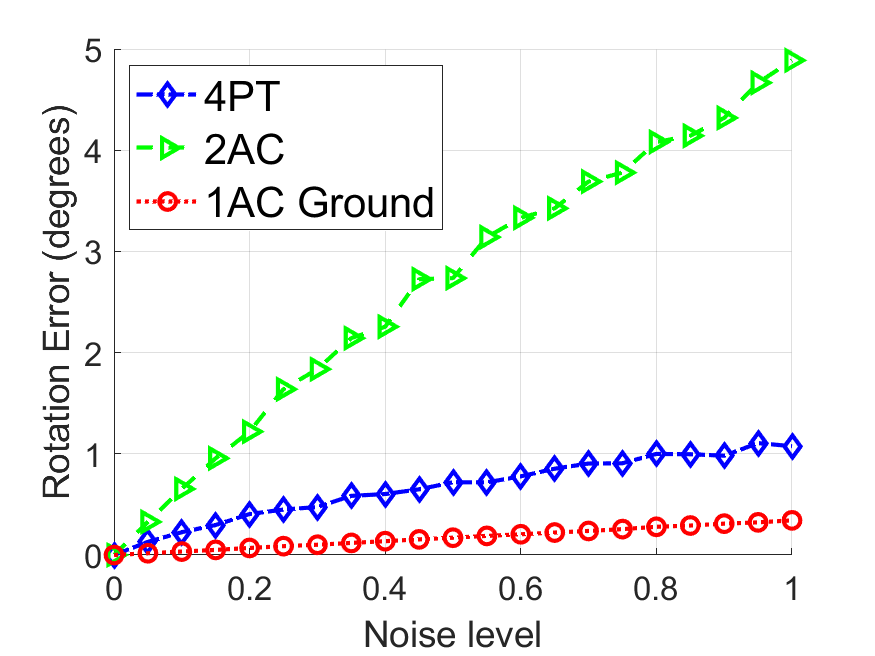}
   	\caption{Forward motion; ground plane.}
   	\label{fig:syn_groundPlane_rotError}
	\end{subfigure}
	\begin{subfigure}[h]{0.30\textwidth}
	    \centering
    	    \includegraphics[width=\columnwidth]{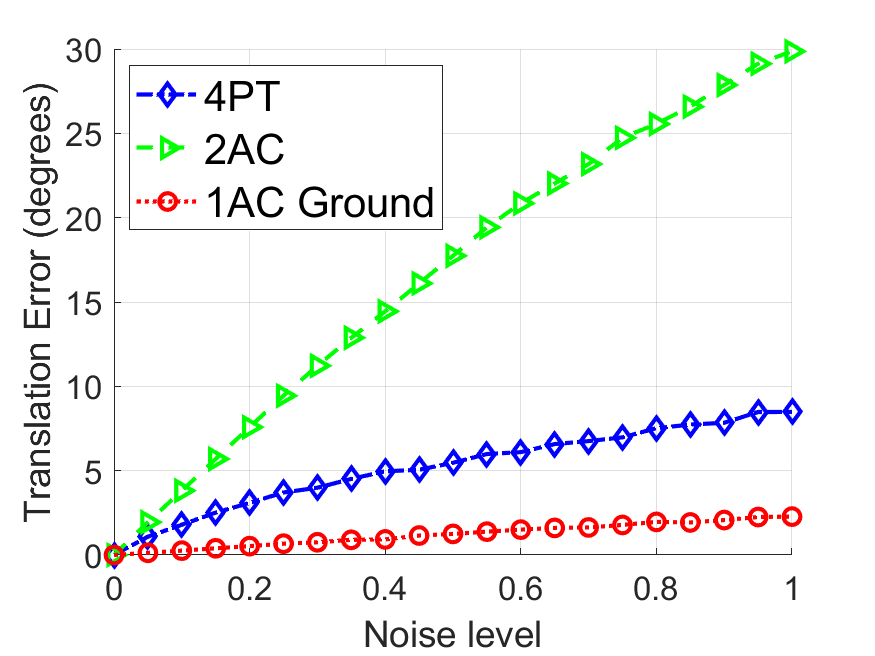}
	\caption{Forward motion; ground plane.}
	\label{fig:syn_groundPlane_transError}
    \end{subfigure}
    \begin{subfigure}[h]{0.30\textwidth}
	    \centering
    	    \includegraphics[width=\columnwidth]{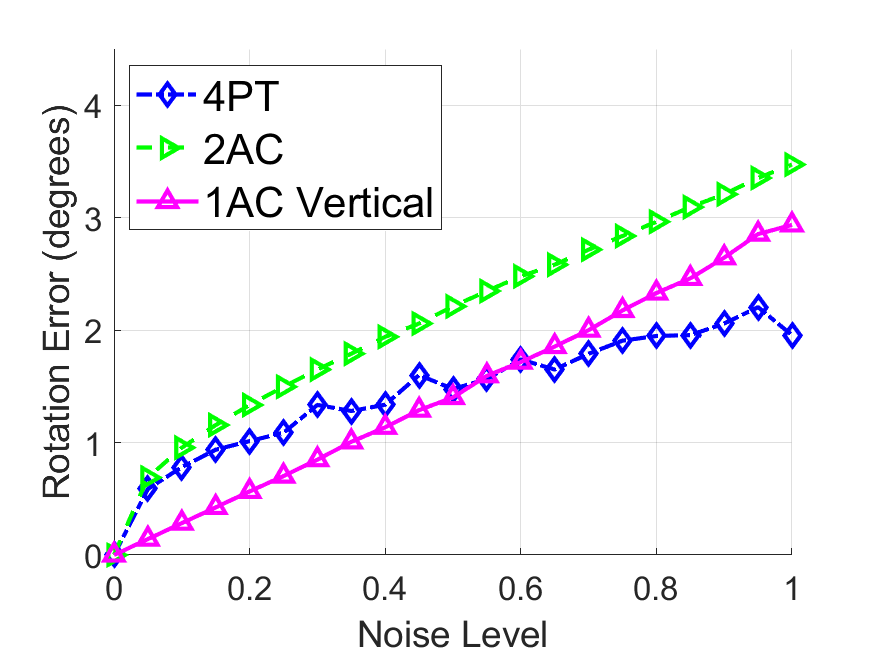}
	\caption{Forward motion; frontal plane.}
	\label{fig:syn_frontWall_rotError}
    \end{subfigure}
    
    \begin{subfigure}[h]{0.30\textwidth}
	    \centering
    	    \includegraphics[width=\columnwidth]{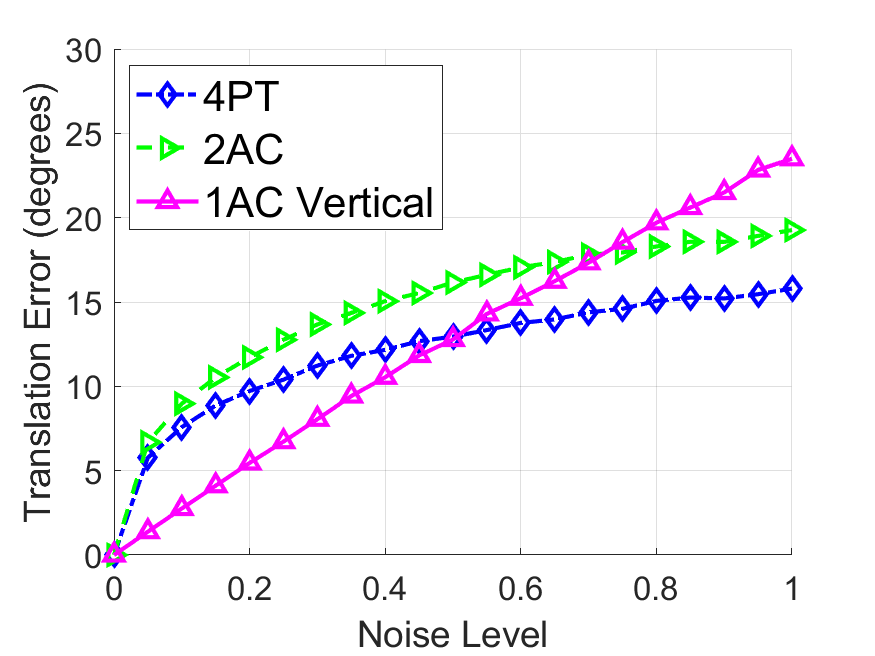}
	   	\caption{Forward motion; frontal plane.}
	   	\label{fig:syn_frontWall_transError}
    \end{subfigure} 
	\begin{subfigure}[h]{0.30\textwidth}
	    \centering
	\includegraphics[width=\columnwidth]{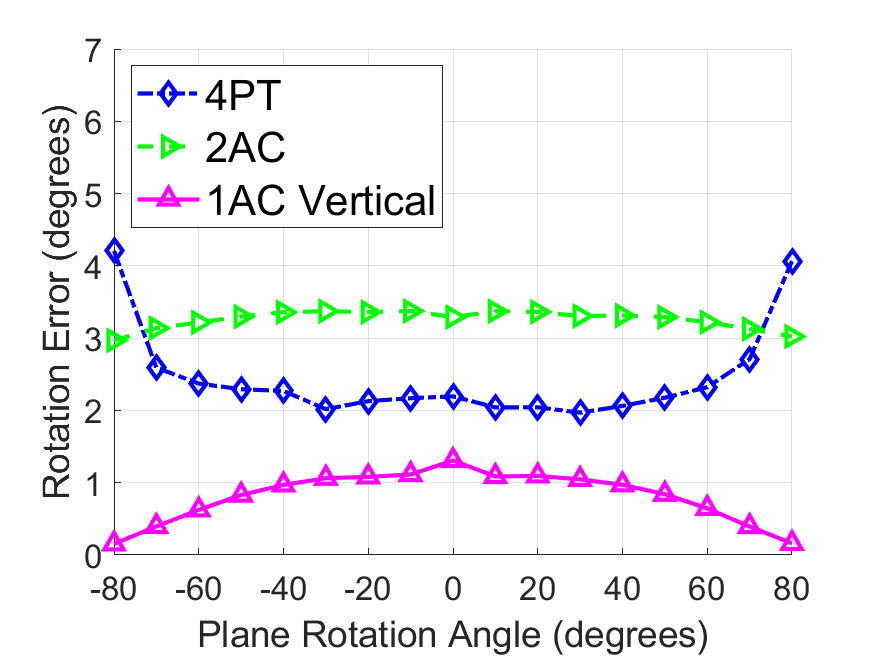}
   	\caption{Forward motion; general vertical plane.}
   	\label{fig:syn_generalWall_rotError_forward}
    \end{subfigure}
    \begin{subfigure}[h]{0.30\textwidth}
	    \centering
    	    \includegraphics[width=\columnwidth]{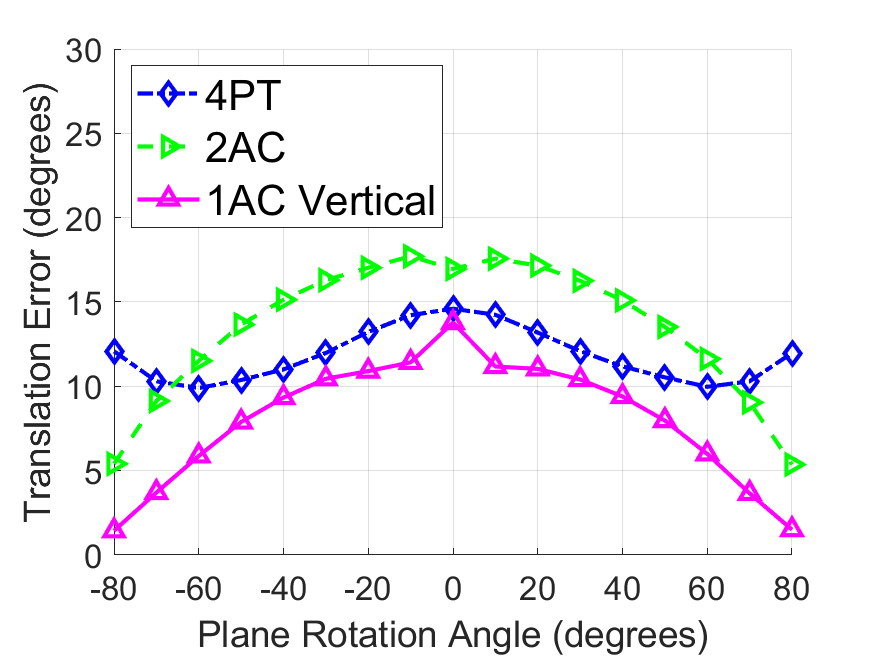}
	\caption{Forward motion; general vertical plane.}
	\label{fig:syn_generalWall_transError_forward}
    \end{subfigure}
    
    \begin{subfigure}[h]{0.30\textwidth}
	    \centering
    	    \includegraphics[width=\columnwidth]{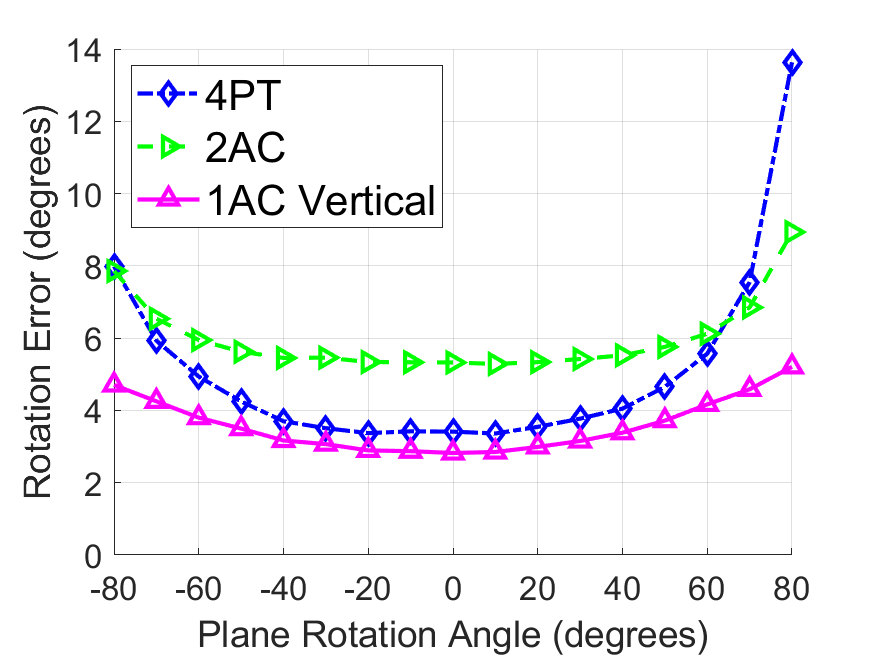}
    \caption{Sideways motion; general vertical plane.}
    \label{fig:syn_generalWall_rotError_sideways}
    \end{subfigure}
	\begin{subfigure}[h]{0.30\textwidth}
	    \centering
	\includegraphics[width=\columnwidth]{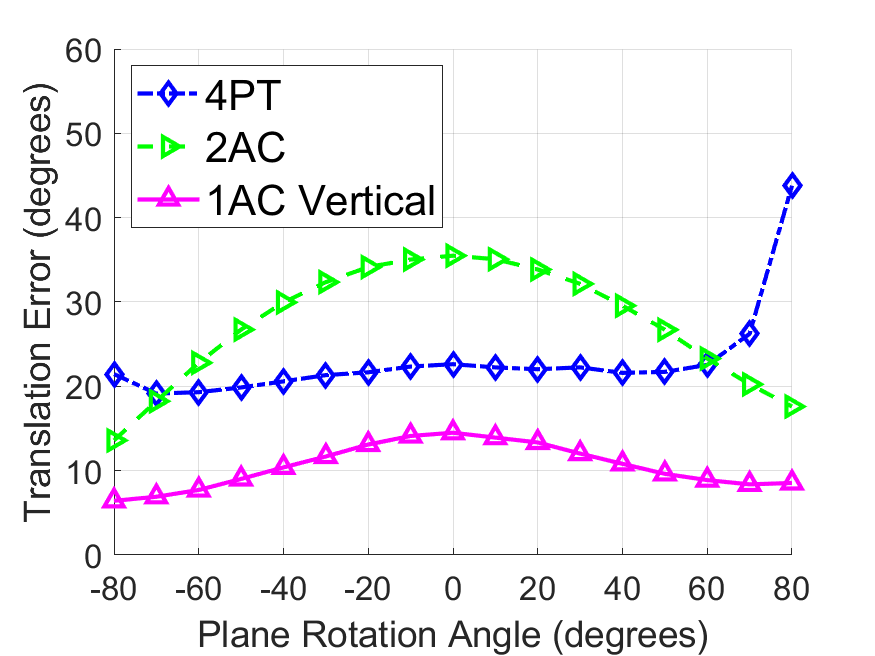}
	\caption{Sideways motion; general vertical plane.}
	\label{fig:syn_generalWall_transError_sideways}
    \end{subfigure}
    \begin{subfigure}[h]{0.30\textwidth}
	    \centering
    	    \includegraphics[width=\columnwidth]{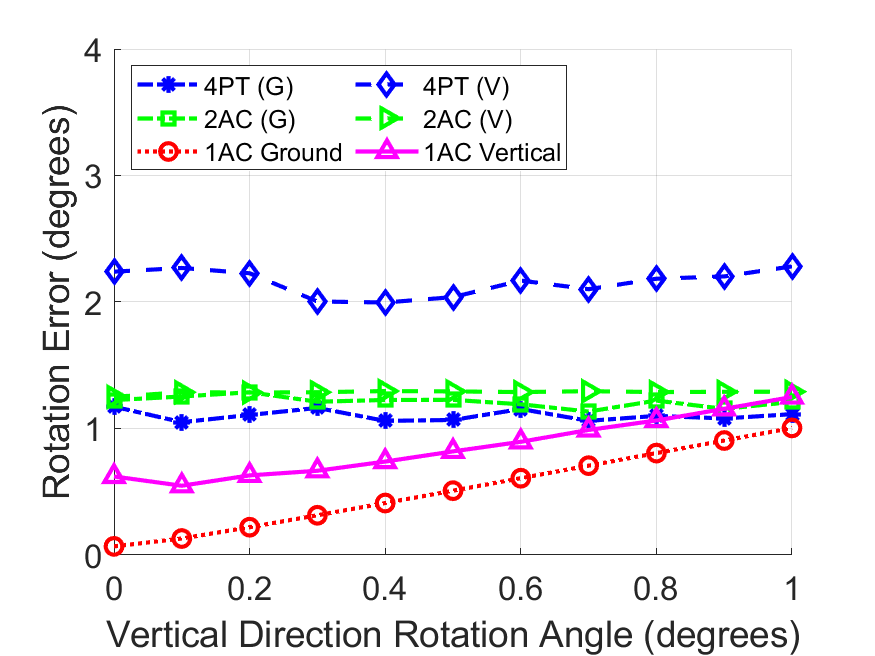}
	\caption{Forward motion; noisy ground and frontal planes}
	\label{fig:syn_gravity}
    \end{subfigure}
    
	\caption{Results on synthetic data. 
	The compared methods: 
	camera pose from an affine correspondence on the ground (1AC Ground); 
	on a frontal wall or on a general vertical plane (1AC Vertical);  
	the normalized DLT~\cite{hartley2003multiple} algorithm (4PC); 
	and the 2AC method~\cite{barath2017theory} estimating general homographies.
	It is shown how the results get affected by noise if the camera undergoes 
	forward ((a)--(f) and (i)) or sideways ((g)-(h)) motion.}
	\label{fig:syn_results}
\end{figure*}

The proposed methods are tested on both synthetic and real-world data. Synthetic tests were generated by Matlab, real-world images were downloaded from Malaga dataset~\cite{Malaga}. 
More results are submitted in the supplementary material.

\subsection{Synthetic Evaluation}

To evaluate the algorithm on synthetic data, we created a similar testing setup as in Saurer et al.~\cite{saurer2016homography}. The scene contains a ground plane, a wall in the front, a wall on the side and walls with general orientation. The distance of the planes from the first camera center was set to one unit distance (u.d.). The baseline between the two cameras was set to $0.1$ unit. The focal length of the camera was set to $1000$ u.d. Each algorithm was evaluated under varying image noise.
All the algorithms were tested under two types of movement, purely forward (along axis $\text{Z}$) and sideways movement (along axis $\text{X}$).

\noindent\textbf{General remarks.} To evaluate the accuracy of the algorithms, we compare the estimated relative pose. Experiments showing the accuracy of the plane normals are in the supplementary material.
To measure the error in the relative rotation between the two cameras, we compute the angular difference between the ground truth and estimated rotations. Since the translation vector is known up to an unknown scale, the error of the translation is the angular difference between the ground truth and estimated translations. 
We calculated the error in the estimated normals only for scenes consisting of walls, since the algorithm for ground plane detection assumes known plane normal. The different relative pose errors are calculated as follows:
\begin{eqnarray*}
\begin{array}{c}
\xi_R = \arccos((trace(\matr R  \dot{ \matr R}^\trans)-1)/2), \\
\xi_t = \arccos((\matr t^\trans \dot{ \matr t})/(||\matr t||~ ||\dot{\matr t}||)), \\
\xi_n = \arccos((\matr n^\trans \dot{ \matr n})/(||\matr n||~ ||\dot{\matr n}||)),
\end{array}
\end{eqnarray*}
where $\xi_R$ is the angular difference in $\matr R$, $\xi_t$ is the  direction similarity in $\matr t$ and $\xi_n$ is the plane normal similarity.
Rotation $\matr R$ and translation $\matr t$ are the ground-truth transformations and $\matr n$ is the plane normal, while $\dot{\matr R},~\dot{\matr t},~\dot{\matr n}$ are the corresponding estimated parameters.

Each data point in the figures represents the average of $10,000$ measurements. In each measurement four plane points are generated.  Correspondences are calculated by projecting the generated plane points to the camera images. Affine transformations are calculated for each correspondence from the ground truth homography matrix~\cite{barath2017theory} and noisy point correspondences. Noise is added both to the projected points and to the corresponding affine transformations. The noise in the point correspondences is realized by adding two dimensional vectors with lengths given in pixels to the points projected to the second camera. To add noise to the affine transformations they are decomposed by SVD. The decomposition of the affine transformation yields two rotation matrices and two singular values. Then two types of noise are added to an affine transformation. 
First, the angle of the rotation matrices are perturbed with a small angle. 
Second, the scale of the affine transformation is changed by slightly perturbing the singular values.

The proposed algorithms are compared to the DLT~\cite{hartley2003multiple} and HA~\cite{barath2017theory} algorithms. DLT and HA are often referred to as 4PC and 2AC, as they require four points and two affine correspondences to estimate the homography parameters, respectively. 
In the synthetic evaluation, we considered three types of simulations. 
In the first one, the $\text{Y}$-axes of the cameras are parallel (Fig.~\ref{fig:syn_groundPlane_rotError}-\ref{fig:syn_frontWall_transError}). 
In this case we increased the noise in the correspondences from $0$ to $1$ pixel, while increasing the rotation and scale error of the affine transformations, respectively, from $0$ to $5$ degrees and $0$ to $5$ percentages. 
In the second simulation (Fig. \ref{fig:syn_generalWall_rotError_forward}-\ref{fig:syn_generalWall_transError_sideways}) the orientation of the vertical plane in the scene is changed while the noises are fixed to 1 pixel, 5 degrees and 5 percentages. 
In the last simulation (Fig. \ref{fig:syn_gravity}) the sensitivity of our algorithms is  tested if the camera motion is not entirely planar. 
For this purpose the vertical direction of the second camera is rotated around axis $X$ with a small degree ranging from 0 to 1 while the noises are fixed to be 1 pixel, 1 degree and 1 percentage respectively.

\noindent\textbf{Ground plane.}
The proposed 1AC-Ground algorithm is compared with 4PC and 2AC in Fig.~\ref{fig:syn_groundPlane_rotError} -~\ref{fig:syn_groundPlane_transError} when the camera undergoes purely forward motion. The proposed method significantly outperforms the other ones even if the data are highly contaminated by noise. 
The results show a similar trend in the case of sideways movement, the corresponding figures are put in the supplementary material. 

\noindent\textbf{Special vertical planes.} For scenes where the observed plane is in the front or on the side, the special cases of the 1AC Vertical algorithm are tested. 
Fig.~\ref{fig:syn_frontWall_rotError} -~\ref{fig:syn_frontWall_transError} compare the special case with plane normal $[0 \quad 0 \quad 1]^\trans$. 
Note that the estimation assuming normal $[1 \quad 0 \quad 0]^\trans$ shows similar behaviour and, therefore, it is in the supplementary material. The results show that the rival algorithms are less accurate when the noise is not too high. For high noise, the point-based method 4PC outperforms the affine-based methods. Remark that both affine and point noises are added to the input data, but the affine-based algorithms (1AC and 2AC) are more sensitive to the affine noise. 

\noindent\textbf{General vertical planes.} Fig. \ref{fig:syn_generalWall_rotError_forward} - \ref{fig:syn_generalWall_transError_sideways} compares the proposed 1AC Vertical algorithm with the 4PC and 2AC algorithms in scenes where a rotated front wall is present. The wall is rotated around axis $\text{Y}$. The angle of rotation ranges from $-80$ to $80$ degrees.
The results of the experiments show that the proposed method is better for relative pose estimation in scenes containing walls with different normals for both sideways and forward movement. In all the cases, the proposed methods outperform the general methods in terms of accuracy of relative pose estimation.

\noindent\textbf{Vertical direction.} In Fig.~\ref{fig:syn_gravity}, the effect of perturbing the vertical direction to slightly invalidate the planar motion assumption is tested. The compared 4PC and 2AC are not affected by this type of noise since they assume general motion between the images. 
The proposed 1AC Ground and Vertical methods are more accurate if the vertical noise is below approx. $1.0$ degree. To the best of our knowledge, the expected error of an IMU is about $0.5$ degree.

\subsection{Real-world experiments}
\begin{figure*}[h!]
  	\centering
  	\includegraphics[width=0.8\textwidth]{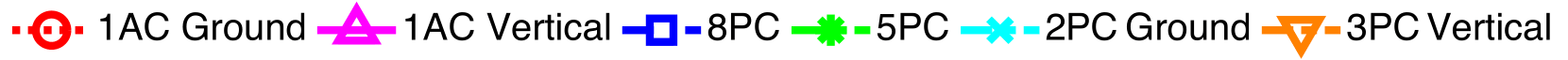}\\[2mm]
  	\includegraphics[width=0.8\columnwidth]{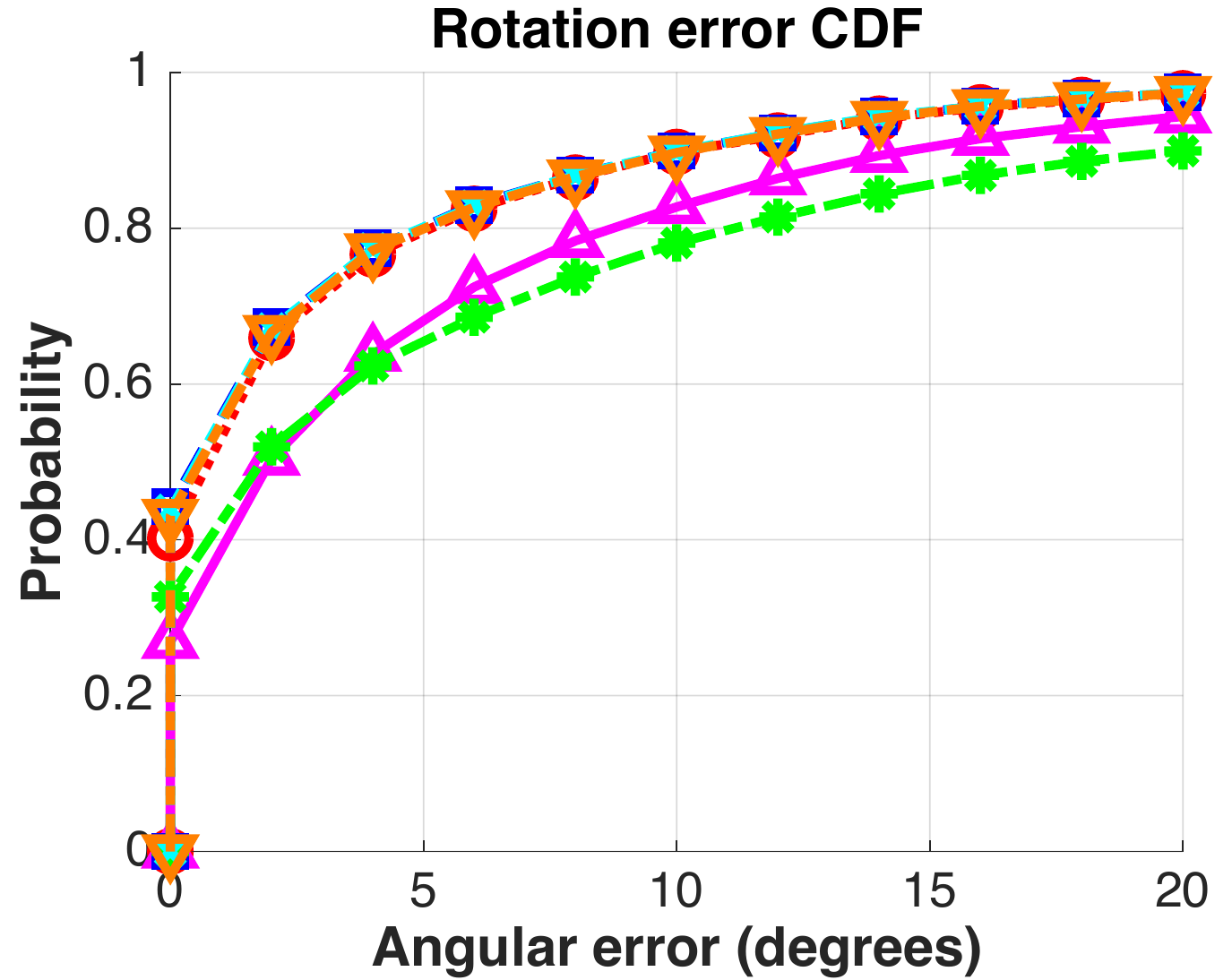}
  	\includegraphics[width=0.8\columnwidth]{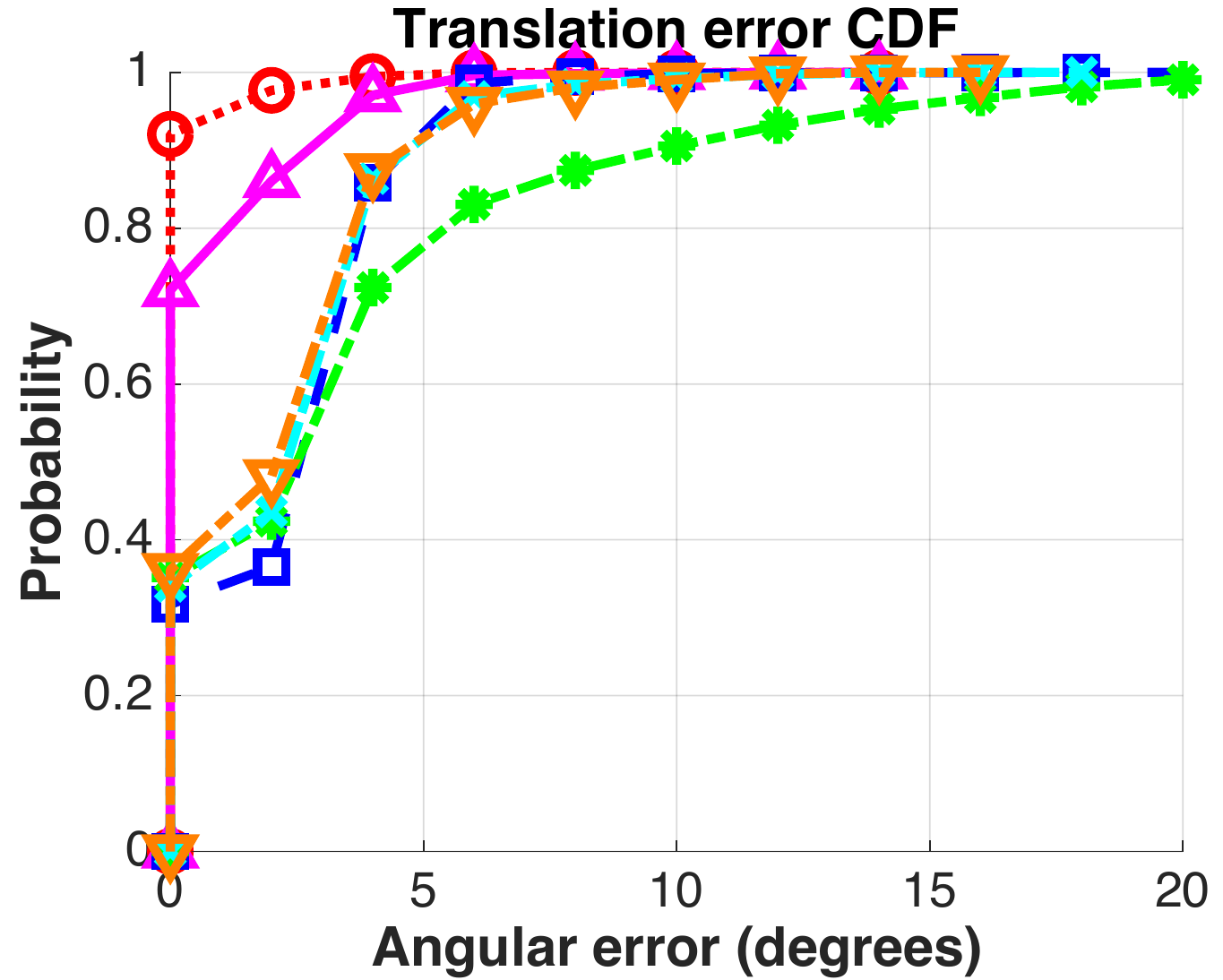}\\[2mm]
  	\includegraphics[width=0.8\columnwidth]{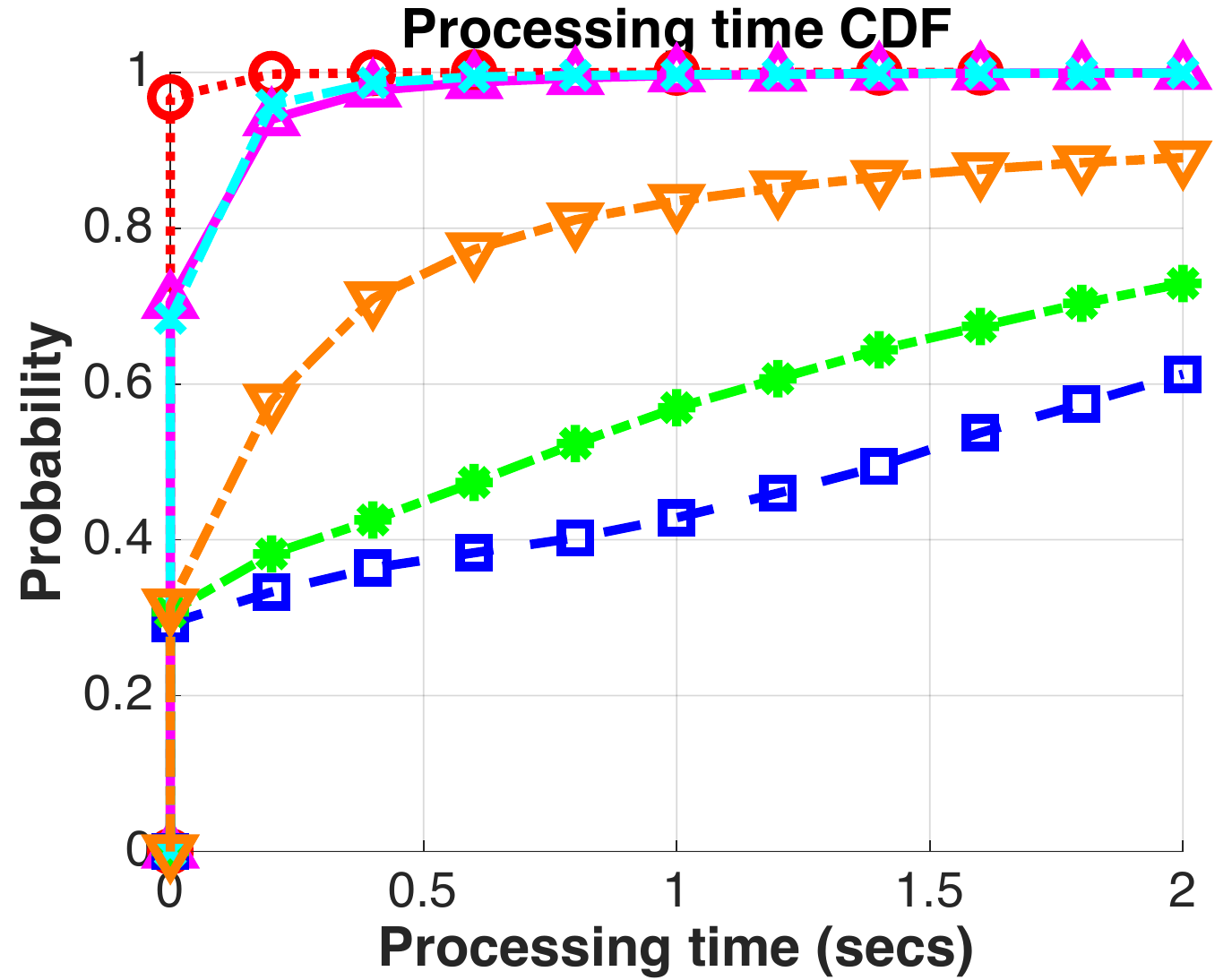}
  	\includegraphics[width=0.8\columnwidth]{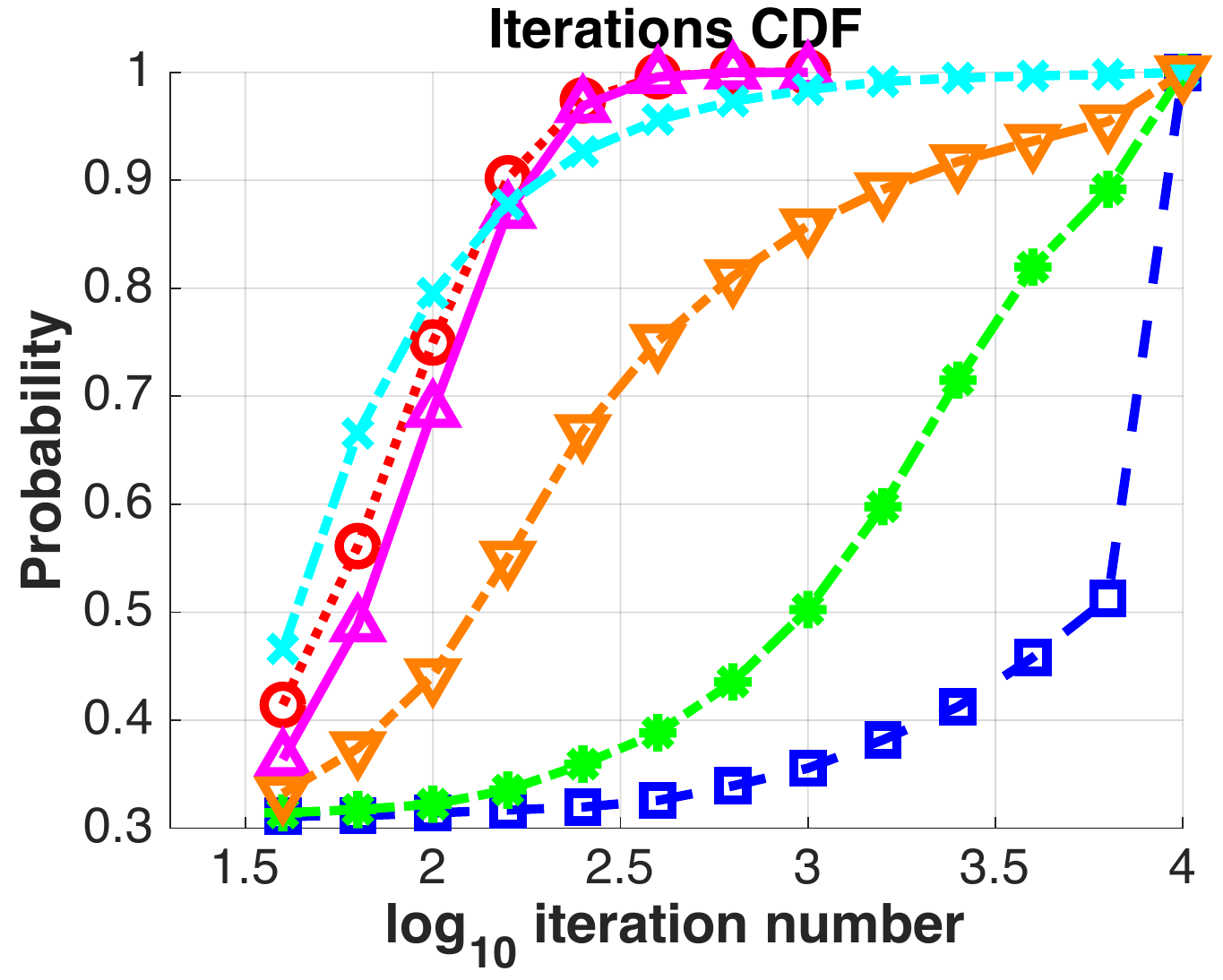}
  	\caption{
  	    The cumulative distribution functions (CDF) of the angular errors (in degrees; top row), processing times (bottom left; in seconds) and $\log_{10}$ number of iterations (bottom right) on the $15$ scenes ($9\;064$ image pairs) of the Malaga dataset are shown. 
  	    Being accurate or fast is interpreted by a curve close to the top-left corner.
  	    GC-RANSAC~\cite{barath2017graph} is used as a robust estimator. 
  	    The compared solvers are the proposed 1AC Ground, 1AC Vertical solvers; the eight (8PC)-~\cite{hartley2003multiple}  and five-point general methods (5PC)~\cite{stewenius2006recent} and the SOTA methods from~\cite{saurer2016homography}, 2PC Ground and 3PC Vertical.}
	\label{fig:real_experiments}
\end{figure*}

\begin{table*}
\centering
\setlength{\tabcolsep}{4pt}
\resizebox{0.8\textwidth}{!}{ \small \begin{tabular}{ | l | r | c c c c c c c c c c c c c c c | c | }
 	\hline
 		\phantom{xx} & & 1 & 2 & 3 & 4 & 5 & 6 & 7 & 8 & 9 & 10 & 11 & 12 & 13 & 14 & 15 & avg. \\
 	\hline
    \parbox[t]{2mm}{\multirow{6}{*}{\rotatebox[origin=c]{90}{Time (ms)}}} & 1AC(G) & \textbf{28} & \textbf{35} & \textbf{11} & \textbf{22} & \textbf{12} & \textbf{17} & \textbf{34} & \textbf{33} & \textbf{26} & \textbf{107} & \textbf{14} & \textbf{34} & \textbf{25} & \textbf{25} & \textbf{33} & \textbf{30}\\
    & 1AC(V) & 136 & 144 & 49 & 86 & 55 & 82 & 178 & 149 & 100 & 417 & 61 & 159 & 102 & 97 & 135 & 130\\
    & 8PC & 4944 & 3291 & 1009 & 2228 & 777 & 3755 & 3480 & 3317 & 2051 & 3494 & 1437 & 2957 & 2734 & 2012 & 1940 & 2628\\
    & 5PC & 3268 & 2761 & 676 & 2785 & 592 & 2717 & 1830 & 1642 & 842 & 3207 & 970 & 1579 & 2113 & 1321 & 1362 & 1844\\
    & 2PC(G) & 168 & 141 & 63 & 98 & 40 & 116 & 139 & 148 & 107 & 228 & 78 & 137 & 135 & 106 & 117 & 121\\
    & 3PC(V) & 1099 & 2190 & 563 & 1356 & 771 & 638 & 2605 & 939 & 456 & 4193 & 437 & 2193 & 1448 & 761 & 1276 & 1395\\
	\hline
    \parbox[t]{2mm}{\multirow{6}{*}{\rotatebox[origin=c]{90}{Ang.\ error ($^\circ$)}}} & 1AC(G) & 3.01 & 5.38 & 2.14 & 2.41 & 4.13 & 5.32 & 4.90 & 3.80 & 7.58 & 10.42 & 3.15 & 4.27 & 8.05 & 4.40 & 4.03 & 4.87\\
    & 1AC(V) & 4.82 & 8.79 & 6.05 & 4.78 & 4.63 & 7.29 & 7.78 & 6.56 & 10.19 & 12.19 & 5.73 & 6.72 & 11.16 & 7.68 & 7.93 & 7.49\\
    & 8PC & 2.98 & \textbf{5.01} & 2.02 & 2.31 & 4.09 & \textbf{5.32} & \textbf{4.83} & \textbf{3.70} & \textbf{6.71} & \textbf{10.02} & \textbf{2.97} & \textbf{4.00} & \textbf{7.83} & \textbf{4.17} & \textbf{3.79} & \textbf{4.65}\\
    & 5PC & 4.45 & 10.85 & 7.38 & 10.37 & 4.57 & 8.35 & 7.03 & 5.40 & 17.60 & 11.42 & 11.29 & 8.64 & 15.55 & 11.42 & 8.16 & 9.50\\
    & 2PC(G) & 3.00 & 5.02 & \textbf{2.01} & 2.37 & \textbf{4.09} & 5.33 & \textbf{4.83} & 3.71 & 6.73 & 10.03 & 3.00 & 4.01 & 8.19 & 4.28 & \textbf{3.79} & 4.69\\
    & 3PC(V) & \textbf{2.98} & 5.21 & 2.04 & \textbf{2.35} & 4.10 & 5.34 & 4.84 & 3.71 & 6.72 & 10.11 & 3.07 & 4.06 & 7.93 & 4.34 & 3.83 & 4.71\\
	\hline
\end{tabular} }
\caption{The avg. run-times (in milliseconds) and rotation errors (in degrees) of relative pose estimation on the 15 scenes (columns) of the Malaga dataset using different minimal solvers and Graph-Cut RANSAC as robust estimator~\cite{barath2017graph} are reported. The compared methods are the five-point solver of Stewenius et al. (5PC)~\cite{stewenius2006recent}, the eight point solver (8PC)~\cite{hartley2003multiple}, two points on the ground (2PC(G)) and three points on a vertical plane (3PC(V)) solvers of Saurer et al.~\cite{saurer2016homography}. and the proposed two affine-based solvers assuming points on the ground (1AC(G)) or on a vertical plane (1AC(V)).
The corresponding cumulative distribution functions are shown in Fig.~\ref{fig:real_experiments}.
}
\label{table:real_results_malaga}
\end{table*}
In order to test the proposed techniques on real-world data, we chose the {\fontfamily{cmtt}\selectfont Malaga}\footnote{\url{https://www.mrpt.org/MalagaUrbanDataset}} dataset~\cite{Malaga}. 
This dataset was gathered entirely in urban scenarios with car-mounted sensors, including one high-resolution stereo camera and five laser scanners. 
We used the sequences of one high-resolution camera and every $10$th frame from each sequence. 
The proposed method was applied to every consecutive image pair. 
The ground truth paths were composed using the GPS coordinates provided in the dataset.
Each consecutive frame-pair was processed independently, therefore, we did not run any optimization minimizing the error on the whole path or detecting loop-closure. The estimated relative poses of the consecutive frames were simply concatenated. 
In total, $9\;064$ image pairs were used in the evaluation.

To acquire affine correspondences~\cite{lowe1999object} we used the VLFeat library~\cite{vedaldi08vlfeat}, applying the Difference-of-Gaussians algorithm combined with the affine shape adaptation procedure as proposed in~\cite{baumberg2000reliable}.
In our experiments, affine shape adaptation had only a small $\sim10$\% extra time demand over regular feature extraction.
The correspondences were filtered by the standard SNN ratio test~\cite{lowe1999object}.

As a robust estimator, we chose Graph-Cut RANSAC~\cite{barath2017graph} (GC-RANSAC) since it is SOTA and its source code is publicly available\footnote{\url{https://github.com/danini/graph-cut-ransac}}.
In GC-RANSAC (and other RANSAC-like methods), two different solvers are used: (a) one for fitting to a minimal sample and (b) one for fitting to a non-minimal sample when doing model polishing on all inliers or in the local optimization step. For (a), the main objective is to solve the problem using as few data points as possible since the processing time depends exponentially on the number of points required for the model estimation. The proposed and compared solvers were included in this part of the robust estimator.
Also, we observed that the considered special planes usually have lower inlier ratio, being localized in the image, compared to general ones. Therefore, we, instead of verifying the homography in the RANSAC loop, composed the essential matrix immediately from the recovered pose parameters and did not use the homography itself. 
For (b), we applied the eight-point relative pose solver to estimate the essential matrix from the larger-than-minimal set of inliers.

In the comparison, we used the SOTA 2PC Ground and 3PC Vertical~\cite{saurer2016homography}, the proposed 1AC Ground and 1AC Vertical algorithms, all estimating special planes and assuming special camera movement. We tested the proposed rapid and optimal solvers on real scenes and found that the difference in the accuracy is balanced by the robust estimator. We thus chose the rapid solver since it leads to no deterioration in the final accuracy, but speeds up the procedure. Note that the solvers are parameterized so that they return the parameters of planes with particular normals. However, when using RANSAC, this effect is balanced by RANSAC via decomposing H to pose and validating the essential matrix. Also, we tested the traditional pose solvers, \ie, the five-point (5PC) and eight-point (8PC) algorithms.

The cumulative distribution functions of the rotation and translation errors (in degrees) of the compared methods are shown in the top row of Fig.~\ref{fig:real_experiments}. 
The error is calculated by decomposing the estimated essential/homography matrices to the camera pose, \ie, 3D rotation and translation.
A method being accurate is interpreted as the curve being close to the top-left corner of the plot.
It can be seen that the 1AC Ground solver leads to similarly accurate rotation matrices as the most accurate solvers. Also, it leads to significantly more accurate translation vectors than the competitor algorithms -- the returned translation vector has $<1$ degree error on the $97\%$ of the tested $9\;064$ image pairs. 
The 1AC Vertical solver also leads to accurate translations, however, its rotation is less accurate than that of the methods, excluding the five-point algorithm.

The cumulative distribution functions of the processing times and iteration numbers of the whole robust estimation procedure are shown in the bottom row of Fig.~\ref{fig:real_experiments}. 
Interestingly, the 1AC Ground method does not always lead to the lowest iteration number (right plot), however, being very efficient, it is far the fastest algorithm (left).
The 1AC Vertical method is the second fastest having similar processing time as the 2PC Ground algorithm.
The average rotation errors and processing times, for each scene, are reported in Table~\ref{table:real_results_malaga}.
It can be seen that the proposed 1AC Ground solver is the fastest one on all scenes by being almost an order of magnitude faster than the second fastest solvers, \ie, 2PC Ground.
The accuracy of methods 1AC Ground, 8PC, 2PC Ground and 3PC Vertical is almost identical with a maximum $0.22$ degrees difference between them.

In summary, the methods are proposed for homography estimation. Thus, our primary objective in the synthetic experiments was to compare them with other homography estimators, e.g., the 2AC and normalized 4PC methods. In the real-world experiments, we considered it as an advantage of the formulation that the relative pose is also estimated via the homography. We, therefore, compare the proposed methods to SOTA relative pose solvers, where the real benefit of the proposed approach lies. The proposed 1AC Ground method leads to the most accurate translation vectors and comparable rotation matrices while being significantly faster than the SOTA algorithms.

\section{Conclusion}

We proposed two minimal solvers for estimating the egomotion of a calibrated camera mounted to a moving vehicle.
It is recovered via estimating the homography from a single affine correspondence.
Each proposed solver assumes that we have a prior knowledge about the plane to be estimated. 
Both methods are solved as a linear system with a $6 \times 5$ coefficient matrix, therefore, being extremely fast, \ie, $5$--$10$ $\mu$s
in C++. 
The methods are tested on synthetic data and on publicly available Malaga dataset. 
The technique solving for the ground plane achieves exceptionally efficient performance with similar accuracy to the SOTA algorithms.


{\small
\bibliographystyle{ieee}
\bibliography{egbib}
}

\end{document}